\documentclass[runningheads]{llncs}
\usepackage{listings}
\usepackage{setspace}
\usepackage{pifont}
\usepackage{multirow}
\usepackage{graphicx,wrapfig,lipsum}
\usepackage{algorithm}
\usepackage{algpseudocode}
\usepackage{subcaption}
\usepackage{geometry}
\usepackage{hyperref}
\usepackage{amssymb}
\usepackage{xcolor}
\usepackage{pifont}
\begin{document}
\title{SALAD: \underline{S}mart \underline{A}I \underline{L}anguage \underline{A}ssistant \underline{D}aily}
%
%
\author{Ragib Amin Nihal\inst{1}\orcidID{0000-0002-3876-3033} \and
Tran Dong Huu Quoc\inst{1} \and
Lin Zirui\inst{1} \and
Xu Yimimg\inst{1} \and
Liu Haoran\inst{1} \and
An Zhaoyi\inst{1} \and
Kyou Ma\inst{1}
}
\authorrunning{Nihal et al.}
%
\institute{Tokyo Institute of Technology}
\maketitle              
\begin{abstract}

SALAD is an AI-driven language-learning application designed to help foreigners learn Japanese. It offers translations in Kanji-Kana-Romaji, speech recognition, translated audio, vocabulary tracking, grammar explanations, and songs generated from newly learned words. The app targets beginners and intermediate learners, aiming to make language acquisition more accessible and enjoyable. SALAD uses daily translations to enhance fluency and comfort in communication with native speakers. The primary objectives include effective Japanese language learning, user engagement, and progress tracking. A survey by us found that $39\%$ of foreigners in Japan face discomfort in conversations with Japanese speakers. Over $60\%$ of foreigners expressed confidence in SALAD's ability to enhance their Japanese language skills. The app uses large language models, speech recognition, and diffusion models to bridge the language gap and foster a more inclusive community in Japan. \href{https://drive.google.com/file/d/1qk0cIsAJVhyWGH_PX1Nu_abB7mGF6aEL/view?usp=drive_link}{\textcolor{blue}{SALAD Demonstration}}.

\end{abstract}
\section{Introduction}
\subsection{Problem Overview}
Foreigners living in Japan often have a hard time learning Japanese. When they talk to each other, they usually use translators like Google Translate \cite{Google} and DeepL \cite{deepl}, but these tools do not give them a full language-learning experience. The problems become clear when there are not any useful features for learning the language and there is nothing fun or interesting about it. According to a survey by us, approximately 39\% of surveyed individuals admitted feeling discomfort during conversations with native speakers.

\subsection{Motivation}
Our motivation for developing SALAD is to enhance language proficiency among learners, enabling them to communicate with native Japanese speakers confidently and comfortably. Recognizing the power of music as both an effective language learning tool and a source of entertainment, we aim to create an engaging and enjoyable educational experience. We seek to offer a personalized, self-paced learning journey that goes beyond the one-size-fits-all approach of common language platforms, tailoring the experience to the unique needs of each user.

\subsection{Objectives}
\begin{enumerate}
    \item \textbf{Integration of Communication, Learning, and Entertainment:} Seamlessly blend features for effective communication facilitation, language learning, and entertainment within a unified platform.
    \item \textbf{Progress Tracking, Customized and Self-Paced Learning}: To offer a personalized learning experience, allowing users to track their progress, and provide dynamic methods like daily translations, speech recognition, and personalized vocabulary tracking to enhance their Japanese language skills.
\end{enumerate}
\section{Survey Discussion}
A survey (Survey Link: \url{https://forms.office.com/r/4CmShdeeFq}) of 41 participants from 15 Japanese nationalities revealed language challenges faced by foreigners. We asked the following questions:
\begin{itemize}
    \item Nationality \& Age
    \item How long have you been in Japan? \& Japanese Language Proficiency
    \item How often do you interact with native Japanese speakers in your daily life?
    \item How comfortable do you feel engaging in casual conversations with Japanese speakers?
\item Do you use translators (Google, DeepL, etc.) to communicate from English (or other languages) to Japanese?
\end{itemize}
\begin{figure}[htbp]
         \centering
         \includegraphics[scale=0.8]{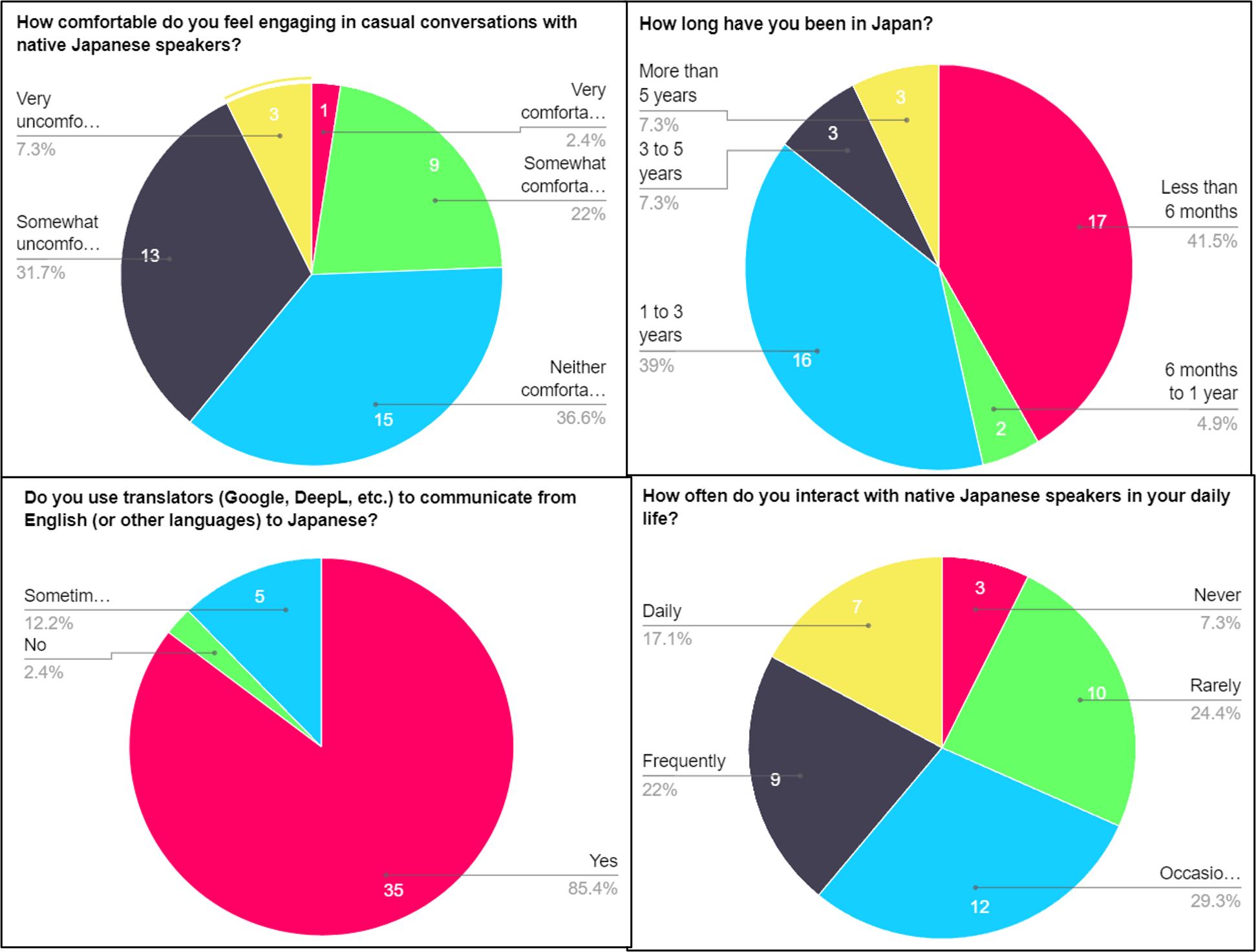}
         \caption{Some of the survey results}
         \label{fig:DNN}
     \end{figure}
From the survey we saw that, 41.5\% have been in Japan for less than six months, indicating varying exposure levels. Communication with native Japanese speakers is varied, with 29.3\% occasionally and 17.1\% daily. However, 31.7\% feel uncomfortable, highlighting communication barriers. 85.4\% rely on translators like Google and DeepL for assistance, highlighting the need for external tools. The survey highlights the need for improved language-learning approaches, and the SALAD app aims to address these needs.

\section{System Design}

\subsection{System Workflow}
The system comprises several main modules: Translation, Vocabulary, Grammar, Lyrics, and Song. Each module serves a distinct function in the final application. The diagram below presents a system architecture that combines different standalone modules.
\begin{figure}[htbp]
         \centering
         \includegraphics[scale=0.8]{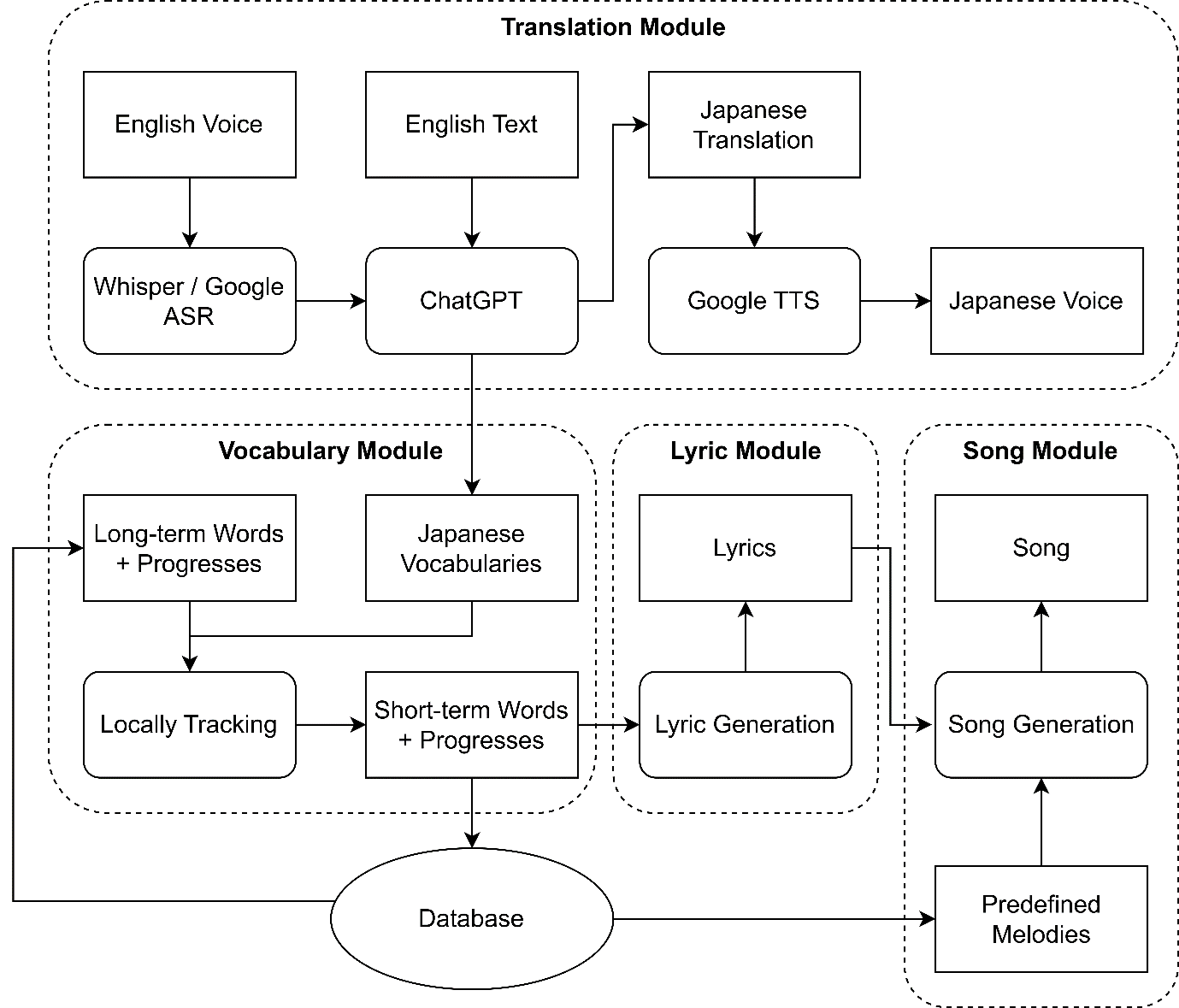}
         \caption{System Architecture: The SALAD system architecture seamlessly integrates four key modules—Translation, Vocabulary, Lyrics, and Song—into a cohesive language learning and musical creation platform. The workflow begins with user input, either spoken or written in English, which is transcribed into text by advanced speech recognition technologies like Whisper or Google ASR. The text is then translated into Japanese using ChatGPT's translation services. ChatGPT also aids in generating vocabulary and grammar analyses, while Google TTS technology converts text into speech for auditory learning. Progress tracking is meticulously managed, with local session data providing immediate feedback and a central database documenting long-term learning. The system's creative aspect emerges in the Lyrics and Song modules, where it crafts music by integrating learning words into pre-existing melodies to produce complete songs. This user-centric design supports both immediate and extended engagement, offering a transformative approach to language and music learning.}
         \label{fig:worflow}
     \end{figure}

This system addresses language learning and musical creativity using advanced technologies such as speech recognition, text-to-speech synthesis, artificial intelligence, and database management. The workflow of this system can be divided into multiple interconnected components, which will be further explained in this section.
\subsubsection{Module 1: Input and Speech Recognition}
SALAD's input module allows users to input text through typing or voice input. The initial step in the workflow involves recording and handling user input, which can be in the form of spoken English or text. When input is spoken, the system utilizes automatic speech recognition technology, such as Whisper \cite{whisper} or Google ASR, to accurately transcribe spoken words into written text. This transcription process is essential, as it allows the system to convert natural speech into a text format that can be processed by our models. For written English input, the system interacts directly with the text, without the need for speech recognition.  This integration enhances user experience, facilitates immediate language translation, and provides real-time feedback for improved translation and learning.

\subsubsection{Module 2: Translation and Grammar}
SALAD leverages the ChatGPT API \cite{chatgpt} on the backend to translate English phrases into Japanese, presenting the output in Kanji, Kana, and Romaji. This process is augmented by prompt engineering, refining the efficiency and accuracy of the translations. To enhance the learning experience, the module also includes a text-to-speech (TTS) feature using the gTTS \cite{gTTS} library, enabling users to listen to the correct pronunciation of the translated sentences. 
\\
Following the translation, the module employs the same ChatGPT API to generate concise yet comprehensive grammar explanations, which are crucial for understanding the structure and nuances of the Japanese language. This integration of advanced language models ensures a robust and intelligent learning aid for users.
\begin{figure}[htbp]
         \centering
         \includegraphics[scale=0.5]{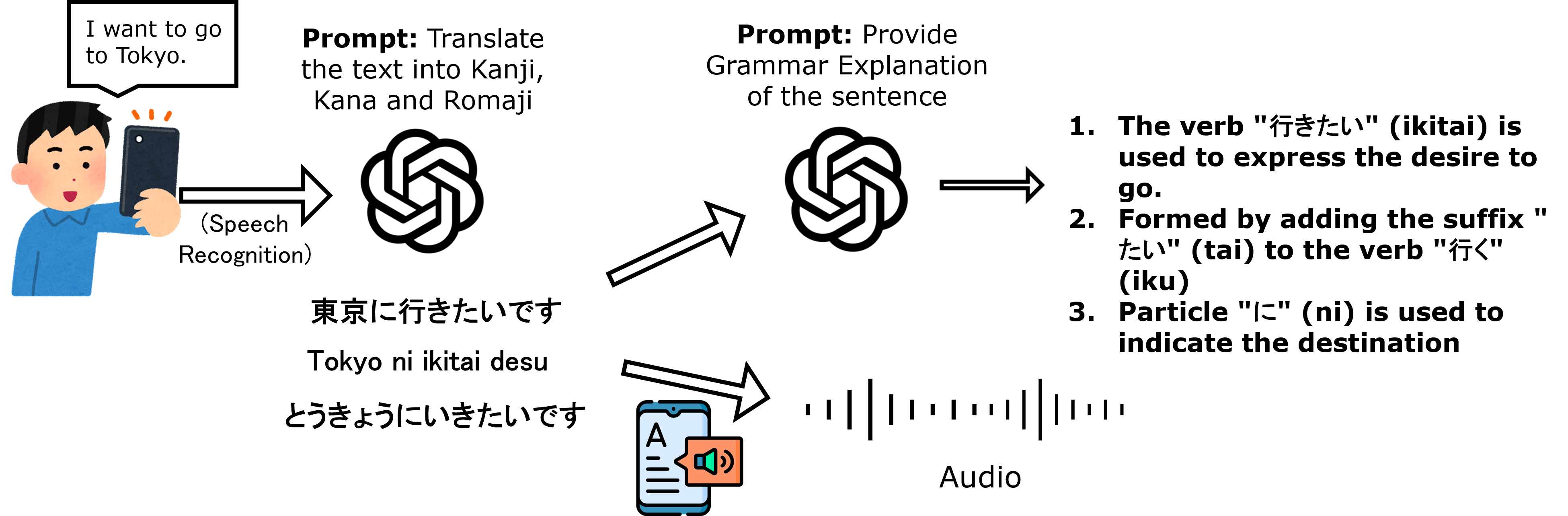}
         \caption{Example of Module 1 and 2}
         \label{fig:DNN}
     \end{figure}
\subsubsection{Module 3: Vocabulary Tracking}
Module 3 in the SALAD system is devoted to tracking and enhancing the user's Japanese vocabulary. It operates by taking the Kanji translation from the user's input or the system's output and processing it through the ChatGPT API with a tailored prompt designed to isolate and identify new vocabulary words. The system then cross-references these words against a database of words the user has already learned, effectively distinguishing between known and new vocabulary. Newly identified words are presented to the user along with their progress and meanings, creating a dynamic and personalized learning experience. The module employs a progress tracking system where a word is considered fully learned once it reaches a progress value of 5, indicating the user's growing proficiency and retention of the Japanese vocabulary. The algorithm is presented below: 
\begin{algorithm}
\caption{Vocabulary Tracking Algorithm}\label{alg:vocab_tracking}
\begin{algorithmic}[1]
\Procedure{TrackVocabulary}{$translated\_sentence$, $database$}
    \For{each $word$ in $translated\_sentence$}
        \If{$word$ is in $database$}
            \State $progress \gets database[word].progress$
            \If{$progress < 5$}
                \State $database[word].progress \gets progress + 1$
            \EndIf
        \Else
            \State $database[word].progress \gets 1$
            \State $database[word].meaning \gets$ \Call{GetMeaning}{$word$}
        \EndIf
    \EndFor
    \For{each $word$, $data$ in $database$}
        \State \Call{DisplayWordProgress}{$word$, $data.meaning$, $data.progress$}
    \EndFor
    \State \Return $database$
\EndProcedure
\\ 
\Function{GetMeaning}{$word$}
    \State $meaning \gets$ \Call{ChatGPT}{$"Define " + word$}
    \State \Return $meaning$
\EndFunction
\\
\Function{DisplayWordProgress}{$word$, $meaning$, $progress$}
    \State Output $word + ": " + meaning + " (Progress: " + progress + "/5)"$
\EndFunction
\end{algorithmic}
\end{algorithm}

\subsubsection{Module 4: Lyrics and Song Generation}
The SALAD system's Lyrics and Song Generation modules combine language learning with musical creativity. They embed user vocabulary into static lyrics, creating personalized lyrics based on user input, AI-generated content, or both. The system uses phoneme units to create songs with varying timbres, allowing for automatic lyrics generation. The singing voice synthesis (SVS) systems within SALAD synthesize high-quality singing voices using DiffSinger \cite{song}, an acoustic model based on the diffusion probabilistic model. A shallow diffusion mechanism is employed to enhance voice quality and expedite inference. The modules are user-centric, supporting both immediate enjoyment and long-term language mastery. They represent a transformative approach to language acquisition, allowing users to engage with the language through song. The success of these modules outperforms state-of-the-art SVS works and is generalizable to other tasks like text-to-speech synthesis. This integration of language and music provides immersive and enjoyable language learning experiences.

\subsection{System Application and User Interface}
Once the system is developed, we merge it with the user interfaces (UI) and export them into a single application. The systems are then linked to two distinct UI designs: one designed using Gradio \cite{gradio}, and the other using QT6, with the help of PySide6, a Python library that facilitates the development of QT-based GUI applications. Processes are depicted by figure below:

\begin{figure}[htbp]
         \centering
         \includegraphics[scale=0.7]{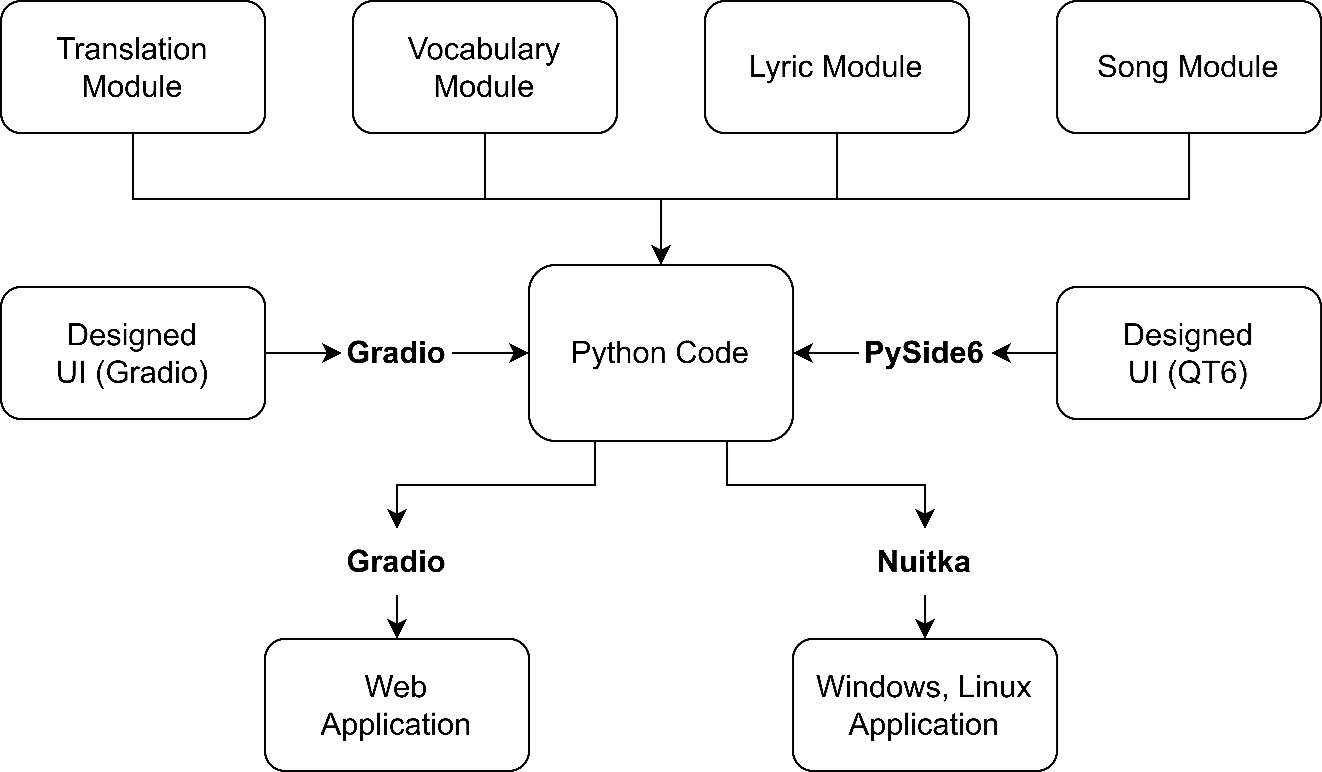}
         \caption{System Integration in User Interface}
         \label{fig:DNN}
     \end{figure}

\paragraph*{Web Application Development with Gradio:}

\begin{itemize}
    \item \textbf{Interface Encapsulation:}
    Define callable interfaces for Translation, Vocabulary, Lyric, and Song modules within the Python codebase.
    
    \item \textbf{Web Interface Design:}
    Use Gradio to create a web-based interface that connects with the Python backend.
    
    \item \textbf{User Interaction:}
    Enable users to interact with the system via a web browser, with Gradio managing the input and output flow.
    
    \item \textbf{Deployment:}
    Deploy the web application, allowing users to access the system's functionalities through Gradio's execution of the Python code.
\end{itemize}

\paragraph*{Desktop Application Development with PySide6 and Nuitka:}

\begin{itemize}
    \item \textbf{UI Connection:}
    Employ PySide6 to link the QT6-designed user interface with the Python code, ensuring responsive desktop UI interactions.
    
    \item \textbf{Compilation:}
    Compile the integrated Python application using Nuitka, converting the code into C for enhanced performance.
    
    \item \textbf{Executable Creation:}
    Generate a self-contained executable for Windows and Linux platforms, encapsulating all modules and UI elements.
    
    \item \textbf{Distribution:}
    Distribute the desktop application as a standalone product, providing an alternative to the web application.
\end{itemize}

\paragraph*{Final Outcome:}

The system is available in two formats: a web application accessible through Gradio and a desktop application compatible with Windows and Linux. Both versions maintain the same essential features, offering users flexibility in how they choose to engage with the application. This dual-platform approach ensures a broad reach and caters to varying user preferences, ultimately delivering a comprehensive and versatile language learning and music creation tool.

\section{System Implementation}
The video demonstration of the system can be found here: \href{https://drive.google.com/file/d/1qk0cIsAJVhyWGH_PX1Nu_abB7mGF6aEL/view?usp=drive_link}{\textcolor{blue}{SALAD Demonstration}}.

\begin{figure}[htbp]
  \centering

  \begin{subfigure}[b]{0.44\textwidth}
    \includegraphics[width=\textwidth]{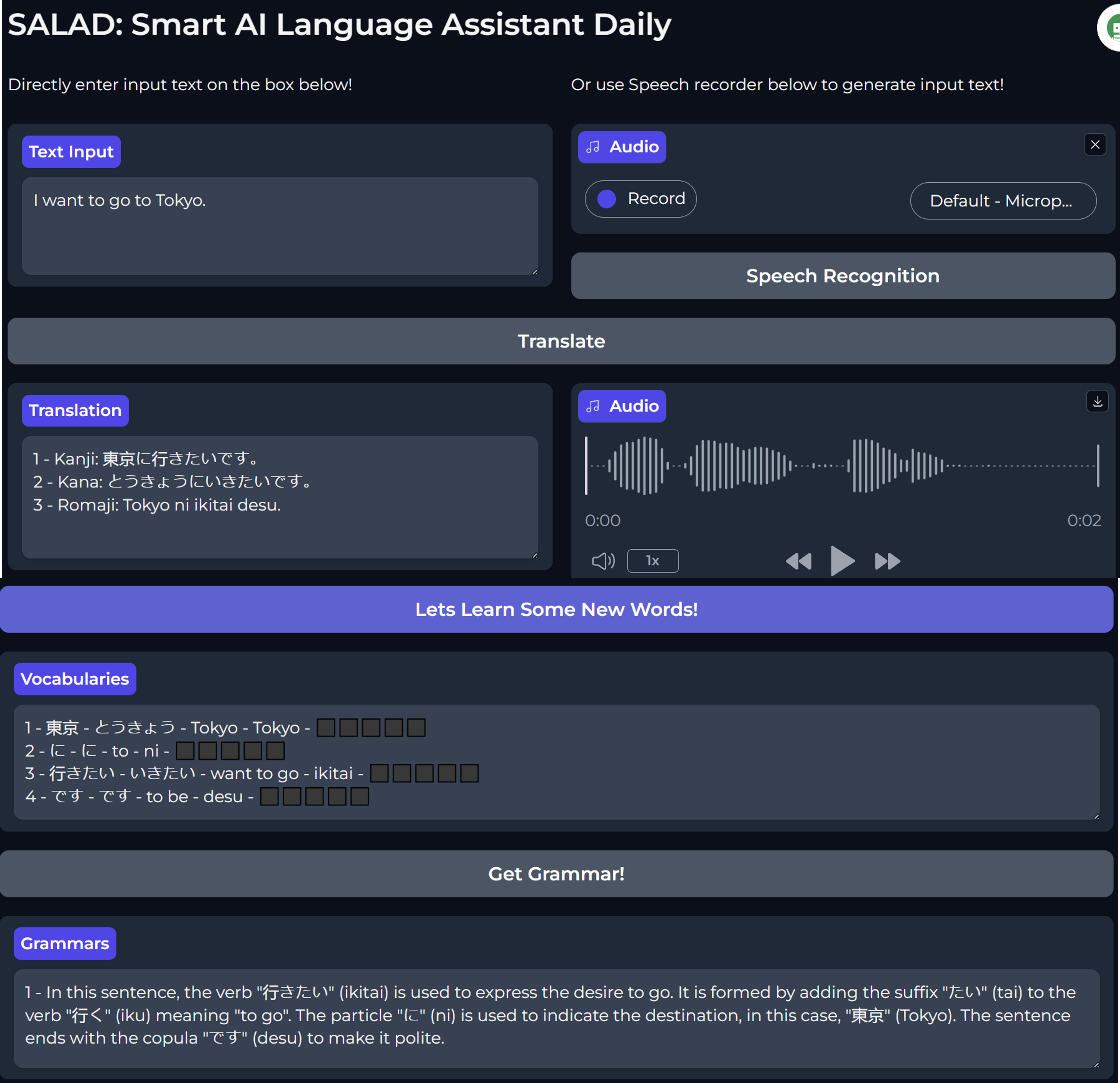}
    \caption{Web Application of SALAD}
    \label{fig:sub1}
  \end{subfigure}
  \hfill
  \begin{subfigure}[b]{0.53\textwidth}
    \includegraphics[width=\textwidth]{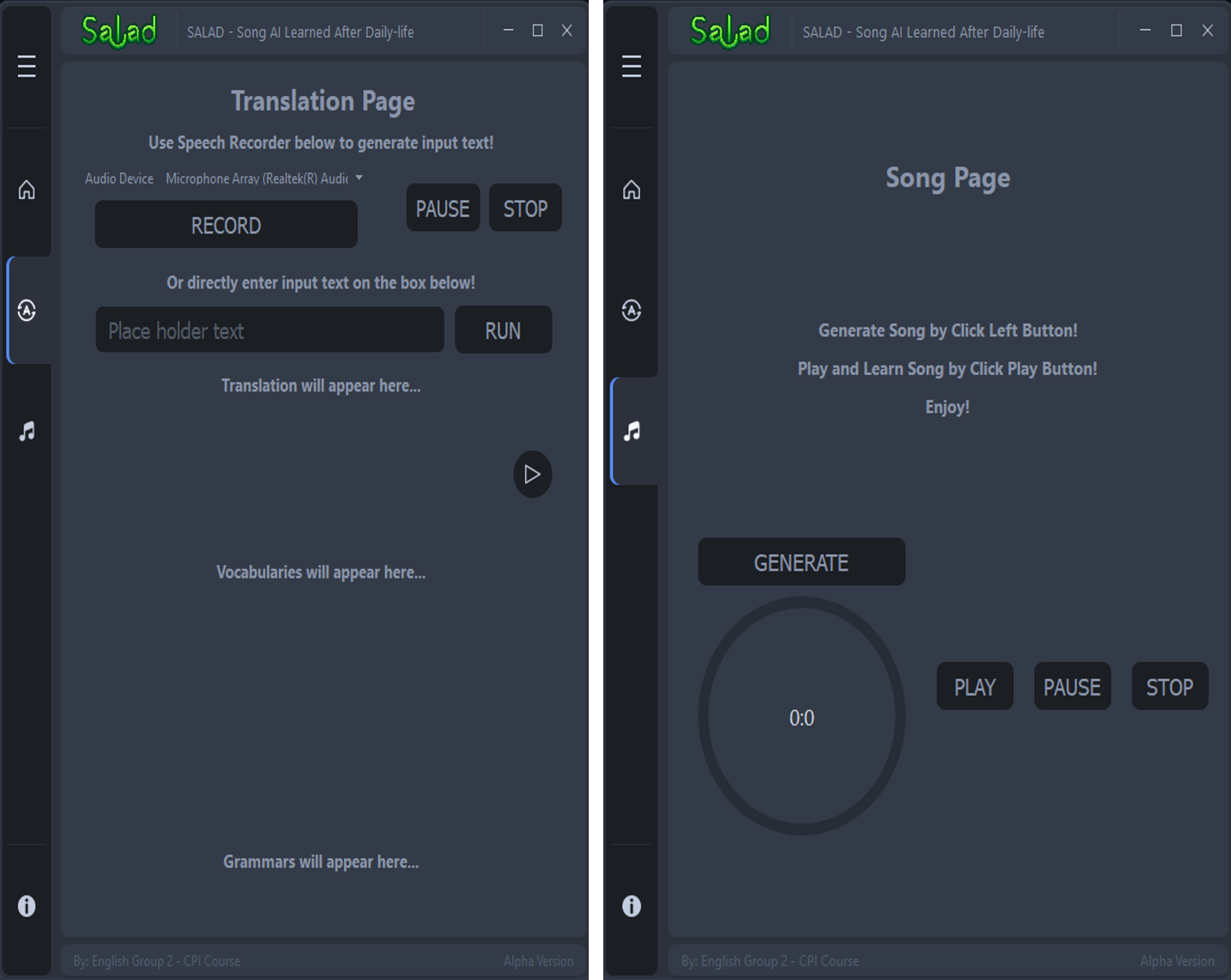}
    \caption{Desktop Application of SALAD}
    \label{fig:sub2}
  \end{subfigure}

  \caption{System Implementation in UI}
  \label{fig:main}
\end{figure}

Figure \ref{fig:sub1} is the web application developed by Gradio UI. there are two options available to users for translation. They can either enter English text in the left box or capture English voice in the right box. In the case of the right box, once the voice is recorded and the Speech Recognition button is clicked, the translated text will be shown in the left box. Each button is associated with a particular module, and clicking on them will display the output of each module in the box below.
\\
Simillary in Figure \ref{fig:sub2}, the QT6 UI is shown. Users can use the recording function to capture audio. The text resulting from automatic speech recognition (ASR) will be displayed in the text box below. By clicking the RUN button, all outputs, including translated Japanese text, vocabulary, and grammatical information, will be presented and filled in the table. With the song generation user interface, users can click the GENERATE button to obtain learning songs. Once the song generation module is finished, the duration will appear on the circle, and users can click the PLAY button to start learning.

\section{Discussion}

\subsection{User Impression}
We also took survey of the user's impression of the app. Below some of the results are shown:
\begin{figure}[htbp]
         \centering
         \includegraphics[scale=0.7]{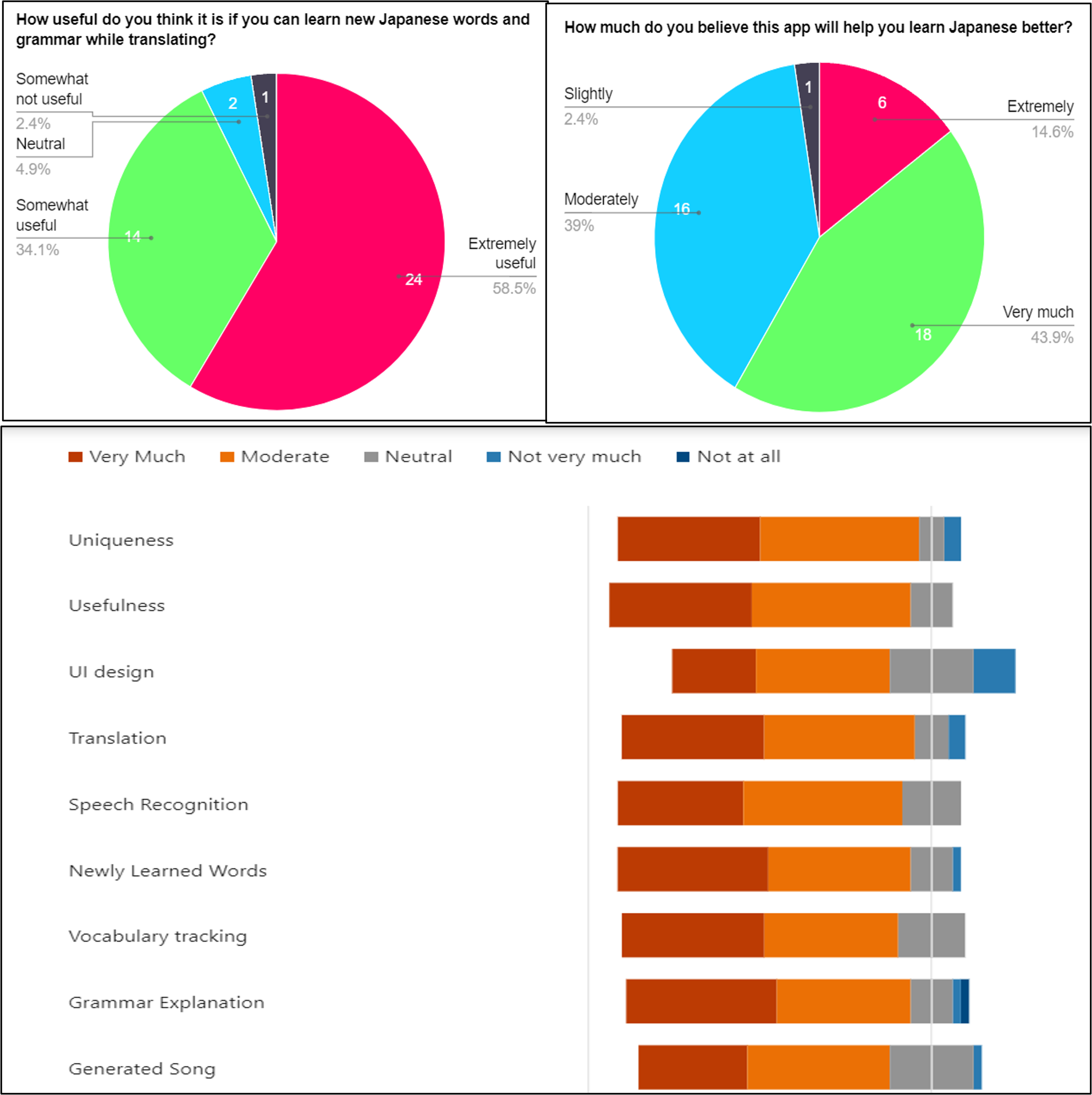}
         \caption{User Impression Survey}
         \label{fig:DNN}
     \end{figure} 

The survey conducted to assess user impressions of the SALAD app's features has yielded overwhelmingly positive responses. A significant majority of users, at 58.5\%, find the ability to learn new Japanese words and grammar while translating to be ``Extremely useful," with an additional 34.1\% rating it as ``Somewhat useful." This indicates a strong user belief in the efficacy of integrated language learning tools, suggesting that the application's approach aligns well with the users' learning preferences.

Moreover, when asked about the app's potential impact on their Japanese language proficiency, a combined total of 58.5\% of users expressed high confidence, with 14.6\% expecting the app to be ``Extremely" helpful and 43.9\% anticipating it to be ``Very much" beneficial. This optimism is further supported by the 39\% who believe the app will be "Moderately" helpful, underscoring the users' overall positive expectations of the app's capabilities in enhancing their language skills.

The absence of any users believing that the app would be ``Not at all" useful, along with the minimal percentage considering it only ``Slightly" or ``Somewhat not useful," underscores the strong user belief in the value of the SALAD app as an effective tool for learning Japanese. These impressions highlight the importance of multifaceted language learning applications that not only provide translation services but also facilitate a deeper understanding and practical knowledge of the language.
\subsection{Comparison with Other Translators}
The comparison table illustrates that while traditional translators excel in translation accuracy, the SALAD App stands out by seamlessly integrating language learning and entertainment. With features like speech recognition, vocabulary building, and grammar explanation, SALAD offers a comprehensive and engaging language-learning experience, addressing the limitations of conventional translators.
\begin{table}[htbp]
\centering
\begin{tabular}{|c|c|c|}
\hline
\textbf{Feature} & \textbf{Traditional Translators} & \textbf{SALAD App} \\
\hline
Translation & \textcolor{green}{\checkmark} & \textcolor{green}{\checkmark} \\\hline
Speech Recognition & \textcolor{green}{\checkmark} & \textcolor{green}{\checkmark} \\\hline
Language Learning Integration & \textcolor{red}{\ding{55}} & \textcolor{green}{\checkmark} \\\hline
Vocabulary Progress Tracking & \textcolor{red}{\ding{55}} & \textcolor{green}{\checkmark} \\\hline
Grammar Explanation & \textcolor{red}{\ding{55}} & \textcolor{green}{\checkmark} \\\hline
Entertainment Value & \textcolor{red}{\ding{55}} & \textcolor{green}{\checkmark} \\\hline
Customization and Self-Paced Learning & \textcolor{red}{\ding{55}} & \textcolor{green}{\checkmark} \\\hline

\end{tabular}
\end{table}

\subsection{Limitations}
While the SALAD system offers innovative solutions for language learning through translation and music, it has certain limitations that need to be addressed. 
\begin{itemize}
    \item Currently, the system is tailored specifically for English to Japanese translation, which limits its applicability to speakers of these two languages. There is potential to expand its scope to include additional language pairs, thereby broadening its user base and utility.
    \item The system's heavy reliance on the ChatGPT API also presents a limitation. While this API provides powerful language processing capabilities, the system's performance is contingent on the availability and functionality of this third-party service. Developing a custom fine-tuned language model (LLM) on the backend could offer more control and potentially enhance the system's capabilities.
    \item In terms of musical output, the system has achieved only partial success in song generation. This area of the application requires further research and development to improve the quality and variety of the generated songs, ensuring that they meet user expectations and effectively aid in language learning.
\end{itemize}

By addressing these challenges, the SALAD system can evolve into a more robust and versatile platform for language acquisition and musical expression.

\section{Conclusion}
The SALAD system represents a significant stride in the realm of language learning, blending the practicality of translation with the joy of music. While the system currently faces limitations such as language specificity and dependency on external APIs, its potential for expansion and refinement is clear. The partial success in song generation points to a promising direction for further enhancement, which could lead to more immersive and effective learning experiences.
As we continue to refine and develop such cyber-physical innovations, we stand on the cusp of a new era in educational technology—one where the barriers of language are not just overcome but are transformed into bridges that connect cultures, countries, and people. We believe thet with persistent effort and ingenuity, systems like SALAD have the potential to significantly reduce communication gaps, fostering greater understanding and unity in our increasingly interconnected world.
%
%
%
%

\end{document}